\documentclass{ifacconf}

\usepackage{graphicx}      
\usepackage{natbib}        
    \usepackage{booktabs}
\usepackage[T1]{fontenc}
\usepackage{subcaption}
\usepackage{graphicx}
\begin{document}
\begin{frontmatter}

\title{Exploring LLM Capabilities for Situational Understanding and COLREG compliance on real-world maritime navigation scenarios} 


\author[DTU]{Julius Wirbel}
\author[DTU]{P. Nicholas Hansen}
\author[DTU,KU]{Line K. H. Clemmensen}
\author[DTU]{Roberto Galeazzi}

\address[DTU]{Technical University of Denmark, Kongens Lyngby, Denmark (e-mail: \{jfmwi, pnha, roga\}@dtu.dk).}
\address[KU]{University of Copenhagen, Copenhagen, Denmark (e-mail: \{lkhc\}@math.ku.dk).}

\begin{abstract}
Recently, Large Language Models (LLMs) have shown considerable capability for situational understanding, reasoning, and decision making in different domains, most notable in the automotive sector. Therefore, we explore current state-of-the-art LLMs as a tool for maritime navigation, which includes both codified rules in the Collision Regulations (COLREGs) and uncodified best practices summarized in the concept of ``Good Seamanship''. We construct a dataset consisting of 50 diverse, real-world navigation scenarios from AIS data, label scenarios with applicable COLREG rules, recommended actions, and the reasoning for the action. We explore a variety of different LLM architectures and sizes to determine their understanding of maritime navigation tasks as well as evaluate their reasoning capabilities in this domain. The results obtained indicate that the maritime navigation task remains difficult to solve without fine-tuning, even for larger online models.

\end{abstract}

\begin{keyword}
Decision and support in marine systems; AI and embodied-AI in marine systems; Marine system guidance, navigation and control
\end{keyword}

\end{frontmatter}

\section{Introduction}

Large Language Models (LLMs) and related multi-modal models have shown promising capabilities in contextual interpretation, rule-based reasoning, and natural-language explanations. These properties make them attractive as high-level reasoning components for autonomous systems, where decisions must be grounded not only in sensor data but also in domain rules, operator intent, and environmental context. Rather than replacing established perception, guidance and control pipelines, such models may serve as cognitive layers that support situation assessment, decision justification and human-machine interaction.

Maritime navigation is a particularly interesting domain to utilize these capabilities. Safe operation at sea is governed by the International Regulations for Preventing Collisions at Sea (COLREGs), but practical decision-making also depends on contextual judgment commonly referred to as good seamanship. Translating these natural-language rules and practices into robust algorithmic behavior remains difficult, especially in dynamic multi-vessel encounters and geographically constrained or regulated waters. Although low-level guidance and control for autonomous surface vessels are well developed, higher-level decision support remains challenging, particularly when systems must interpret encounter geometry, assess navigational responsibility, and justify maneuver choices in a transparent and reliable manner. These challenges are also relevant beyond maritime autonomy, since AI-powered decision-support tools may help reduce human error in conventional watchkeeping operations, where human factors remain the leading contributor to maritime incidents ~\citep{european2022}.

Recent work in autonomous driving suggests that LLM-based systems can contribute meaningfully to high-level reasoning in rule-constrained environments, supported by emerging multimodal planning frameworks, large-scale datasets, and benchmark-oriented evaluations \citep{wangDriveMLMAligningMultiModal2023,simaDriveLMDrivingGraph2025a,liWOMDReasoningLargeScaleDataset2025,cuiLLM4ADLargeLanguage2025,nvidiaAlpamayoR1BridgingReasoning2025}. By contrast, research in maritime autonomy is still at a much earlier stage.

Existing maritime studies have demonstrated the feasibility of LLM-assisted planning and COLREG-aware decision support, but they have largely focused on simplified simulated encounters with limited environmental complexity \citep{christensen-AICaptain,agyei2025largelanguagemodelbaseddecisionmaking}. It therefore remains underexplored how well current off-the-shelf models can interpret real maritime situations in which encounter geometry, vessel behavior, and geographic constraints interact in more dynamic ways.

This paper addresses that gap by evaluating current online and offline LLMs on real-world maritime navigation scenarios derived from Automatic Identification System (AIS) data. Using a dataset of 50 encounters from Danish coastal waters, we assess model performance in three tasks central to maritime decision support: situation classification, responsibility determination, and maneuver recommendation. Rather than proposing a new autonomy architecture or a comprehensive benchmark, the paper provides an initial exploratory assessment of what current off-the-shelf models can do in realistic maritime settings without task-specific fine-tuning. The study is intended as a starting point for deeper analysis and future experimentation on LLM-based situational understanding and COLREG-consistent reasoning.

\section{Related Work}

\subsection{LLMs for autonomous driving}

The automotive domain has led the integration of LLMs into high-level decision-making for autonomous systems. Recent frameworks such as DriveMLM and DriveLM combine multimodal scene understanding with language-based reasoning to connect perception, prediction, and planning in dynamic traffic environments ~\citep{wangDriveMLMAligningMultiModal2023,simaDriveLMDrivingGraph2025a}. In parallel, large-scale resources such as WOMD-Reasoning have enabled the study of explainable planning on real-world driving data ~\citep{liWOMDReasoningLargeScaleDataset2025}. Broader surveys and benchmark-focused efforts, including LLM4AD and Alpamayo-R1, further show that the field is moving toward systematic evaluation of reasoning quality, instruction following, and rule-aware planning ~\citep{cuiLLM4ADLargeLanguage2025,nvidiaAlpamayoR1BridgingReasoning2025}. Taken together, these developments indicate that LLM-based systems are emerging as a promising foundation for advancing autonomous navigation, offering scalable reasoning, interpretability, and rule-aware decision-making capabilities that justify continued research and deployment efforts.

\subsection{LLMs for autonomous ship navigation}

Comparable research in maritime autonomy remains limited. AI Captain introduced one of the first LLM-driven mission planning systems for autonomous surface vessels, combining language-based planning with a hierarchical guidance, navigation, and control (GNC) architecture ~\citep{christensen-AICaptain}. Another recent study \citep{agyei2025largelanguagemodelbaseddecisionmaking} integrated LLMs into COLREG-aware decision-making pipelines, combining language-based reasoning, fuzzy risk assessment, and low-level control to generate explainable collision-avoidance behavior in simulated encounters. These results are promising, but current demonstrations remain largely restricted to simplified and highly controlled simulated scenarios, such as isolated vessel encounters in open water without additional obstacles. As a result, the field still lacks a good understanding of how off-the-shelf models perform on real-world maritime situations characterized by contextual ambiguity, environmental constraints, and more heterogeneous vessel behavior.



\section{Creating Maritime Navigation Scenarios}\label{sec:4}

\begin{table*}[ht]
\centering
\caption{Ground-Truth Distribution of Scenario Type, Ego Role, and Action ($n=50$)}
\label{tab:gt_distributions}
\begin{tabular}{@{}lr @{\hspace{1cm}} lr @{\hspace{1cm}} lr@{}}
\toprule
\multicolumn{2}{c}{\textbf{Scenario Type}} & \multicolumn{2}{c}{\textbf{Ego Role}} & \multicolumn{2}{c}{\textbf{Action}} \\
\cmidrule(r){1-2} \cmidrule(lr){3-4} \cmidrule(l){5-6}
Category & Count (\%) & Category & Count (\%) & Category & Count (\%) \\
\midrule
crossing & 24 [26] (48.0\%) & stand-on & 22 [22] (44.0\%) & maintain course and monitor & 43 [42] (86.0\%) \\
normal navigation & 10 [7] (20.0\%) & give-way & 17 [19](34.0\%) & alter course to starboard & 4 [5] (8.0\%) \\
head-on & 8 [9] (16.0\%)& unclear & 11 [9] (22.0\%) & slow down & 3 [3] (6.0\%) \\
overtaking & 8 [8] (16.0\%) &  &  &  &  \\
\bottomrule
\end{tabular}
\end{table*}

To evaluate the navigation capabilities of off-the-shelf LLMs, a single-shot approach was chosen: based on AIS data collected from Danish waters in 2020, from which a large number of \emph{scenarios} are extracted, where a scenario is defined as a spatial and temporal slice of the AIS data where an encounter is detected. 
An encounter is defined as the violation of a chosen threshold of Distance at Closest Point of Approach (DCPA) and Time to Closest Point of Approach (TCPA) violation between a pair of vessels.
To evaluate the LLM performance, fifty multi-vessel encounters were randomly extracted. For each extracted scenario one vessel was randomly assigned the role of ego-vessel, and all other vessels are designated as target vessels. For all vessels in the scenario, relevant navigation data is extracted, and based on the position data a rendering similar to the Electronic Chart Display and Information System (ECDIS) is created, which contains water and land masses, buoyage, Traffic Separation Schemes (TSS) and vessel positions. The scenarios contain predominantly samples from the Storebælt region, while some samples stem from the Lillebælt region.

\subsection{Vessel State Data Format}\label{sec:data}
Each scenario contains the following textual information, which is provided in a structured JSON format:
\begin{itemize}
    \item \textbf{Vessel States:} Vessel states include information about the current position, speed over ground and course over ground of the vessel. Furthermore, metadata like the length, width and drought of the vessel are extracted for further context. Each scenario contains one state for each vessel.
    \item \textbf{Positional Relations:} Much of maritime navigation is based on the relative position of other vessels to the ego-vessel itself. These relation include the relative bearing used to determine the encounter type, the Closest Point of Approach (CPA), determining the closest distance between two vessels at a given course and speed, and the Time to Closest Point of Approach (TCPA), providing information about the time to the CPA at current course and speed.
    \item \textbf{Scenario Boundaries:} In order to generate the ECDIS chart renderings, a scenario boundary needs to be defined to limit the displayed area and to provide the correct rendering data.
\end{itemize}

\subsection{Chart Rendering}
Charts offer valuable information about the ships environment and are essential to maritime navigation. For commercial vessels, the ECDIS system is used to display information about sea and land masses, buoyage, navigational hazards and the positions of other vessels. Additional overlays like RADAR information can be added on a real-world bridge. For this study, chart data from OpenSeaMap ~\cite{barlocherOpenSeaMapFreeNautical2013} was used to create the renderings based on the water and land masses as well as the buoyage images and navigational information included by these open-source charts. This rendering is centered on the ego-vessel to reflect the representation of the ECIDS system on a vessel bridge.

The following information is included in the rendering:

\begin{itemize}
    \item \textbf{Navigable Waters:} Shown as light-blue polygons indicating safe areas for vessel movement.
    \item \textbf{Land Masses:} Displayed in green to denote non-navigable zones and assess coastal proximity.
    \item \textbf{Buoyage and Traffic Separation Schemes:} Buoys and separation schemes highlight hazards, safe routes, and aid orientation.
    \item \textbf{Vessel Symbols and Predicted Paths:} Vessels are shown with color-coded icons and predicted six-minute course lines.
    \item \textbf{Range Rings:} Overlaid to provide distance reference within the scenario.
\end{itemize}

\section{Semi-Supervised Labeling using GPT-5}

To accelerate development of the evaluation set and associated prompts, GPT-5 was used to generate \emph{initial structured label proposals}, following the general idea of LLM-assisted dataset annotation in prior work \citep{liWOMDReasoningLargeScaleDataset2025,foutterSpaceLLaVAVisionLanguageModel2025a}. GPT-5 was not treated as the final authority. Instead, it was used in a two-step process consisting of chart description and scenario interpretation, after which all labels were manually reviewed and corrected by a human maritime expert.

\subsection{Chart Descriptions} \label{sec:chartdescriptions}
Maritime charts and the information available from the ECDIS hold a variety of relevant clues about the navigation situation the ego-vessel is in. To extract this information, a Vision-Language-Model (VLM) is instructed to output the information with two top-level keys: \textit{description} and \textit{interpretation}:

\begin{itemize}
    \item \textbf{Description:} Captures spatial relationships between the vessel and key chart elements (e.g., land, buoys, channels, TSS, nearby vessels), using relative orientation (port, starboard, ahead, astern) and qualitative distance (near, medium, far).
    \item \textbf{Interpretation:} Prompts the model to infer situational context, such as proximity to land, other vessels, or operation within constrained areas like channels or TSS, with brief supporting rationale.
\end{itemize}

The prompt explicitly prohibits free-form text or COLREG reference to ensure objective, observation-based descriptions. The structured format enables consistent, verifiable extraction of the requested information, creating the input for subsequent situational understanding and rule-based assessment.


\subsection{Scenario Interpretation} \label{sec:output_format}
The LLM prompt for the scenario evaluation and interpretation is engineered to produce a fully structured overview and interpretation of the navigation scenario at hand.
The output format contains three top-level keys: 

\begin{itemize}
    \item \textbf{Scenario Identification:} The model classifies the scenario type (normal navigation, crossing, overtaking, or head-on) and determines the vessel’s role under COLREGs (stand-on, give-way, or unclear), with a brief rationale for its assessment.
    \item \textbf{Applicable Rules:} The LLM identifies relevant COLREG rules for the scenario and provides a short justification for each.
    \item \textbf{Recommended Action:} The model suggests a maneuver (e.g., maintain course, turn to starboard, reduce speed), justifies the recommendation, assigns a risk level (0–10), and explains the risk assessment.
\end{itemize}

\subsection{Label Creation and Validation Process}
The model was run with the lowest available reasoning-depth setting to encourage concise, observation-grounded outputs and reduce hallucination risk. Cases with a TCPA $>$ 120 min were labeled as \emph{normal navigation}, since such long-horizon encounters were considered too uncertain for encounter-specific labeling.

All GPT-5 outputs were manually reviewed before inclusion in the dataset. Final labels were validated and corrected by the first author, who holds a commercially endorsed RYA Yachtmaster Offshore certification and has more than ten years of practical sailing experience, including extensive navigation in high-traffic waters such as the English Channel, the Baltic Sea around the Kiel Canal, and the \O resund. Corrections addressed factual mistakes, missing chart context, and implausible scenario interpretations. Over the 50 scenarios, one critical misclassification was corrected: a head-on encounter was changed to normal navigation as the vessels are on the same course at similar speed, not on a head-on course like the model predicted. Two other scenarios were changed from crossing to normal navigation as the target vessel crossed behind the ego-vessel. Furthermore, 6 image descriptions were corrected to remove buoyage classified as vessels. These mistakes came from small groups of buoys in close proximity to one another. Table~\ref{tab:gt_distributions} summarizes the resulting label distributions for scenario type, ego role, and recommended action.

\section{Evaluation of Open-Source LLMs}
To evaluate the capability of state-of-the-art open-source LLMs in maritime navigation, the following models are compared on the scenario interpretation task:

\begin{itemize}
    \item OpenAI GPT-5-mini \& GPT-4o
    \item MistralAI Mistral 7B 
    \item Microsoft Phi 4 Mini
    \item Google Gemma 3 1B
\end{itemize}

This selection of models represent a variety of parameter sizes and architecture types, from very large online models to small, resource-efficient model variants that can be run on locally, e.g. on a laptop. This selection offers an insight about how these model architectures and sizes compare on the complex maritime navigation tasks.

\subsection{Prompt Engineering}
To achieve a consistent model output, the prompt has been designed to ensure the model only utilizes information provided through the prompt. In addition to the vessel data and chart descriptions, the prompt contains COLREG Rules 6-19 in plain text as well as descriptions of how to identify different encounter types. These explicit descriptions were necessary since initial testing revealed strong rule hallucination in the smaller models. 

The prompt is structured into an instruction and a data block. The instruction block includes the following:
\begin{enumerate}
    \item \textbf{Role and Task Definition:} This part defines the role of the LLM as a navigator, and gives information about the task the model will be performing. Furthermore, this part contains information about the data the LLM will receive from the data block.
    \item \textbf{Scenario Classification Rules:} This part defines the differences between the scenario types, i.e. what differentiates a crossing from a head-on encounter. This was added as during testing without it, the models tended to make up encounter types and repeatedly miss-interpreting the CPA values between crossing, head-on and overtake encounters.
    \item \textbf{Output Format Definition:} This part specifies the JSON output format for the requested information keys described in Section \ref{sec:output_format}.
    \item \textbf{Guidelines:} This final part contains guidelines forcing the network to only used the information provided in the data block, to only output in the specified JSON format and provide concise reasoning in the correct nautical language.
\end{enumerate}

The data block contains the information extracted from the AIS scenario described in Section~\ref{sec:data}, the chart descriptions introduced in Section~\ref{sec:chartdescriptions} and the COLREG rules 6-19 in plain text form. 
The prompt structure and content are identical for all models; however small differences exist, e.g. in the tokens that differentiate the system prompt holding the instructions from the user prompt having the vessel data. These are caused by the requirements of the model architectures and evaluation backend.

\subsection{Evaluation Backends}
For the three ChatGPT models, the OpenAI API in combination with the OpenAI Python library was used to perform the API calls and extract the results. For the remaining models, the HuggingFace Transformers Library ~\citep{wolfHuggingFacesTransformersStateoftheart2020} in combination with Python was used to load the model weights, tokenize the prompts, perform inference and extract the results. 

\section{Result Comparison}
In line with general trends in LLMs, one would expect that the number of parameters of a given model has a great influence on the performance. At a first glance, the results of the accuracy comparison of these five models do not match this expectation: one would expect that the larger, online models like GPT 5-mini and GPT-4o consistently outperform the smaller, offline models. On the surface, this does not seem to be the case, with GPT-4o especially standing out with low relative Ego Role and Action Recommendation scores. On the other hand, especially Phi4-mini and Gemma3-1B seem to show remarkable performance in the Action Recommendation category, matching or outperforming the online models.

\begin{figure}[t]
\begin{center}
\includegraphics[scale = 0.44]{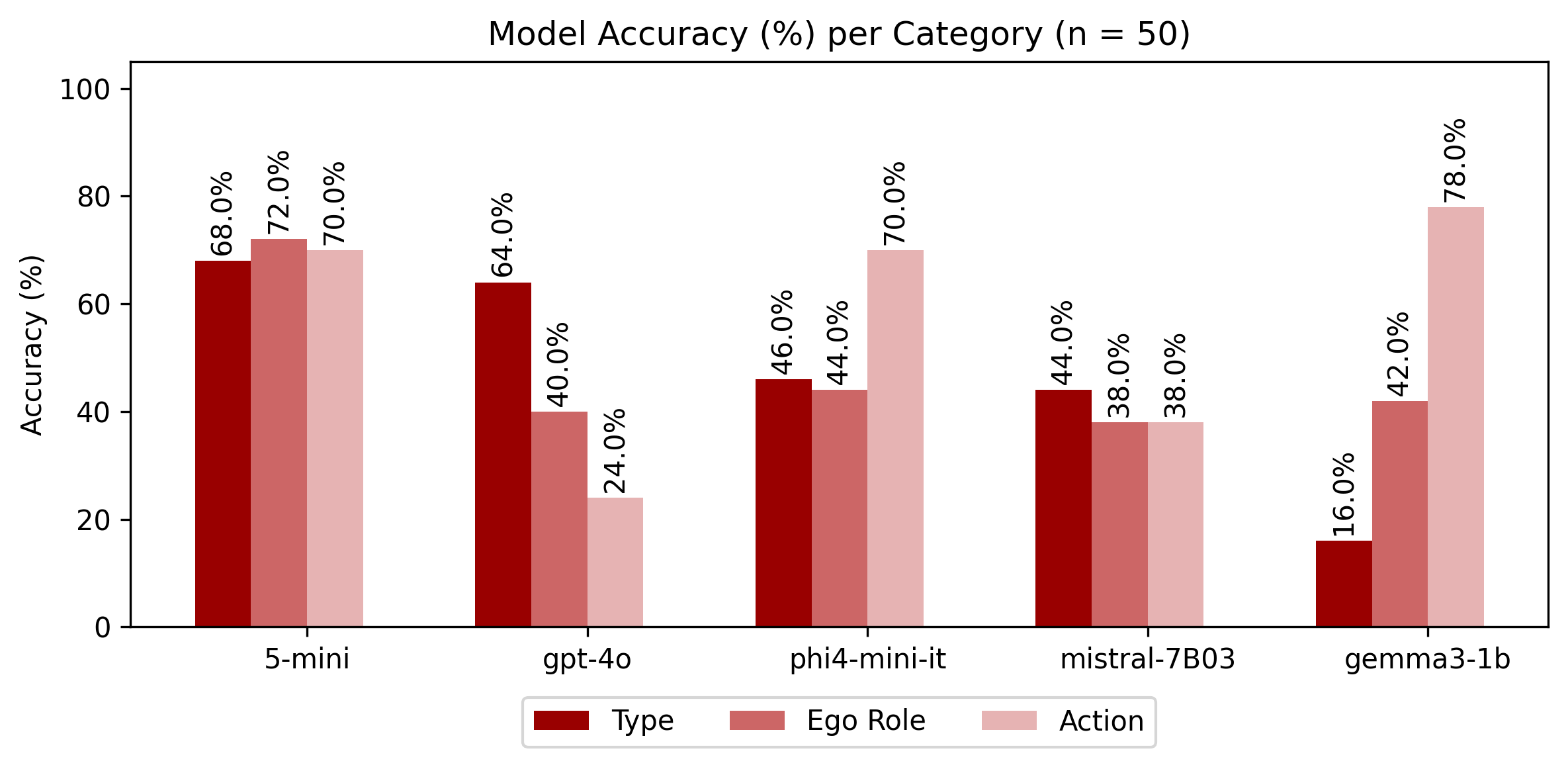}    
\caption{Relative Model Accuracy compared to the baseline.} 
\label{fig:3}
\end{center}
\end{figure}

This focus on accuracy alone could be misleading, and a deeper analysis of the actual model answers for each of the three categories evaluated (see Section \ref{sec:output_format}) is performed.

\begin{figure*}[ht]
\begin{center}
\includegraphics[scale = 0.4]{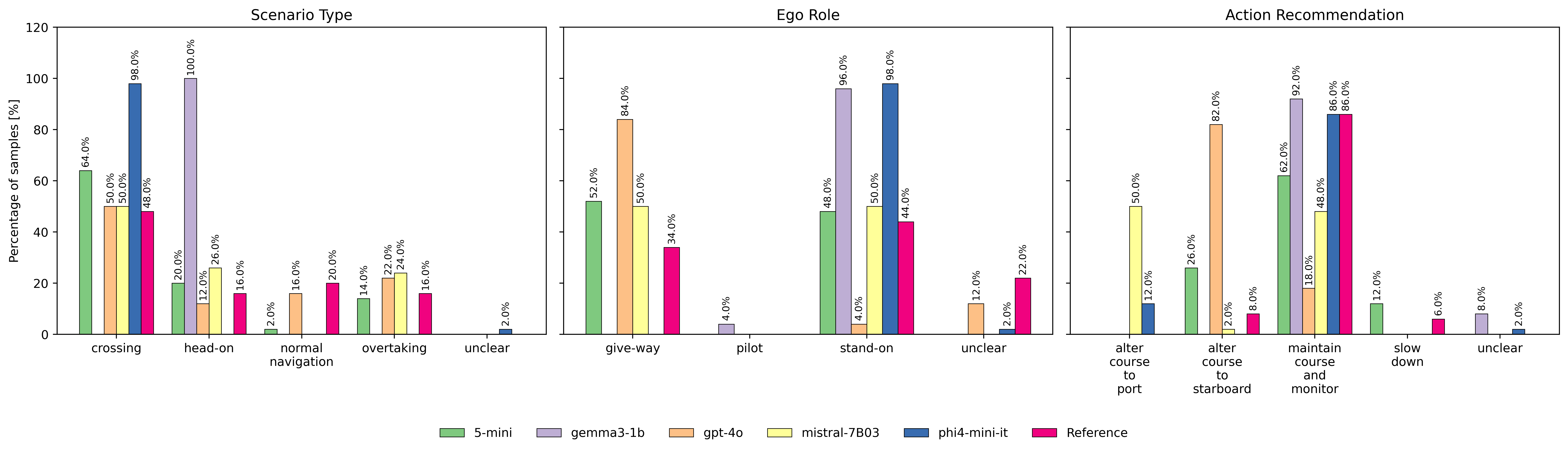}    
\caption{Overview of the relative accuracy for each answer-key combination compared to the baseline.} 
\label{fig:4}
\end{center}
\end{figure*}

\subsection{Scenario Identification}
In the scenario identification task, online models generally align well with the reference labels. GPT-5-mini occasionally confuses overtaking with crossing, while GPT-4o more frequently misclassifies head-on and crossing encounters. Despite high overall accuracy, Phi4-mini and Gemma3-1b fail to generalize, defaulting to a single scenario type regardless of input. Mistral-7B03 appears accurate but struggles with normal navigation and frequently confuses crossing, head-on, and overtaking scenarios. Figure \ref{fig:sub2} provides a detailed breakdown of classification accuracy and misclassification patterns.

\subsection{Ego-Role Identification}
\begin{figure*}[t]
    \centering
    \begin{subfigure}[b]{0.47\textwidth}
        \centering
        \includegraphics[width=\textwidth]{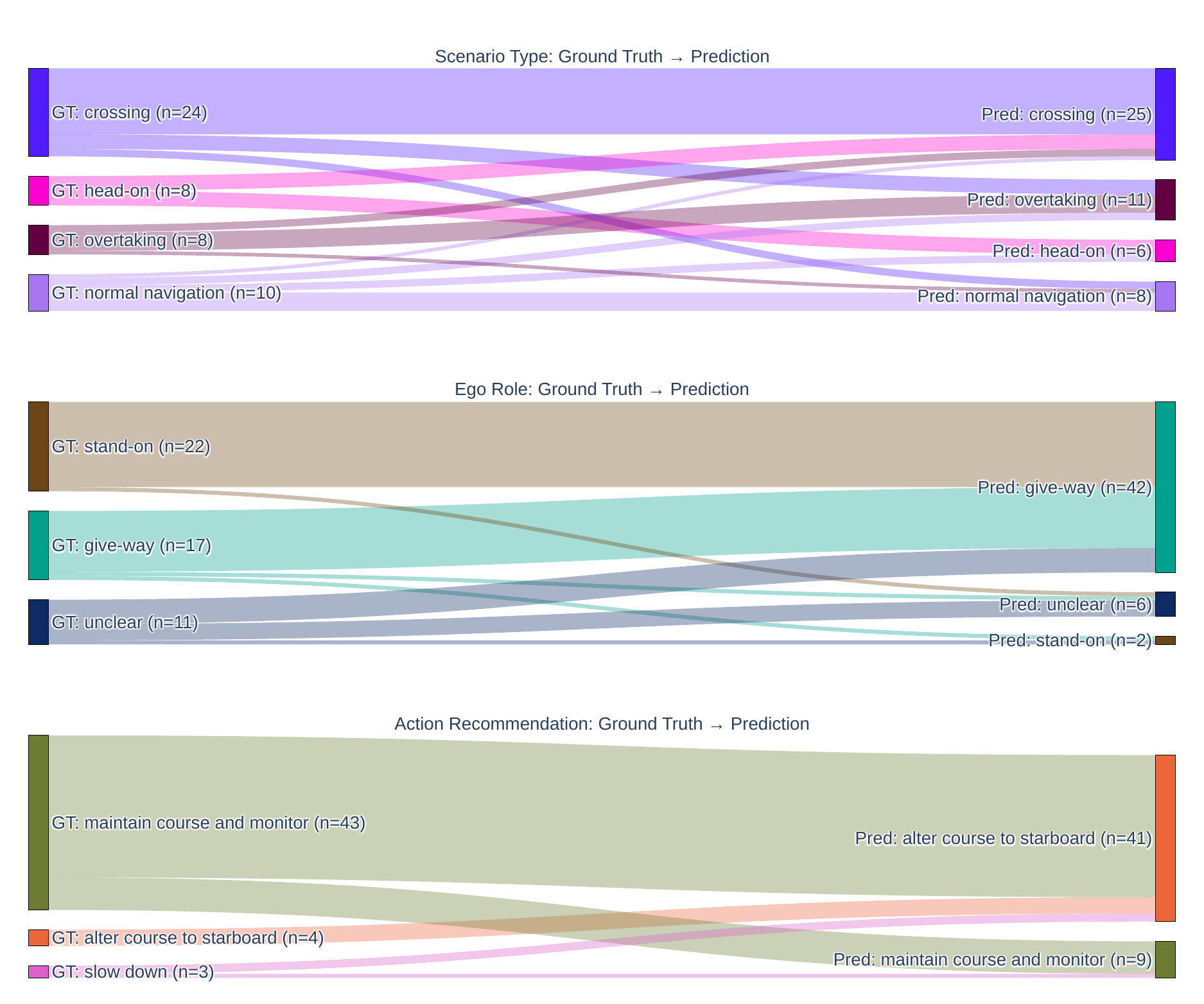}
        \caption{OpenAI GPT-4o}
        \label{fig:sub1}
    \end{subfigure}
    \hfill
    \begin{subfigure}[b]{0.47\textwidth}
        \centering
        \includegraphics[width=\textwidth]{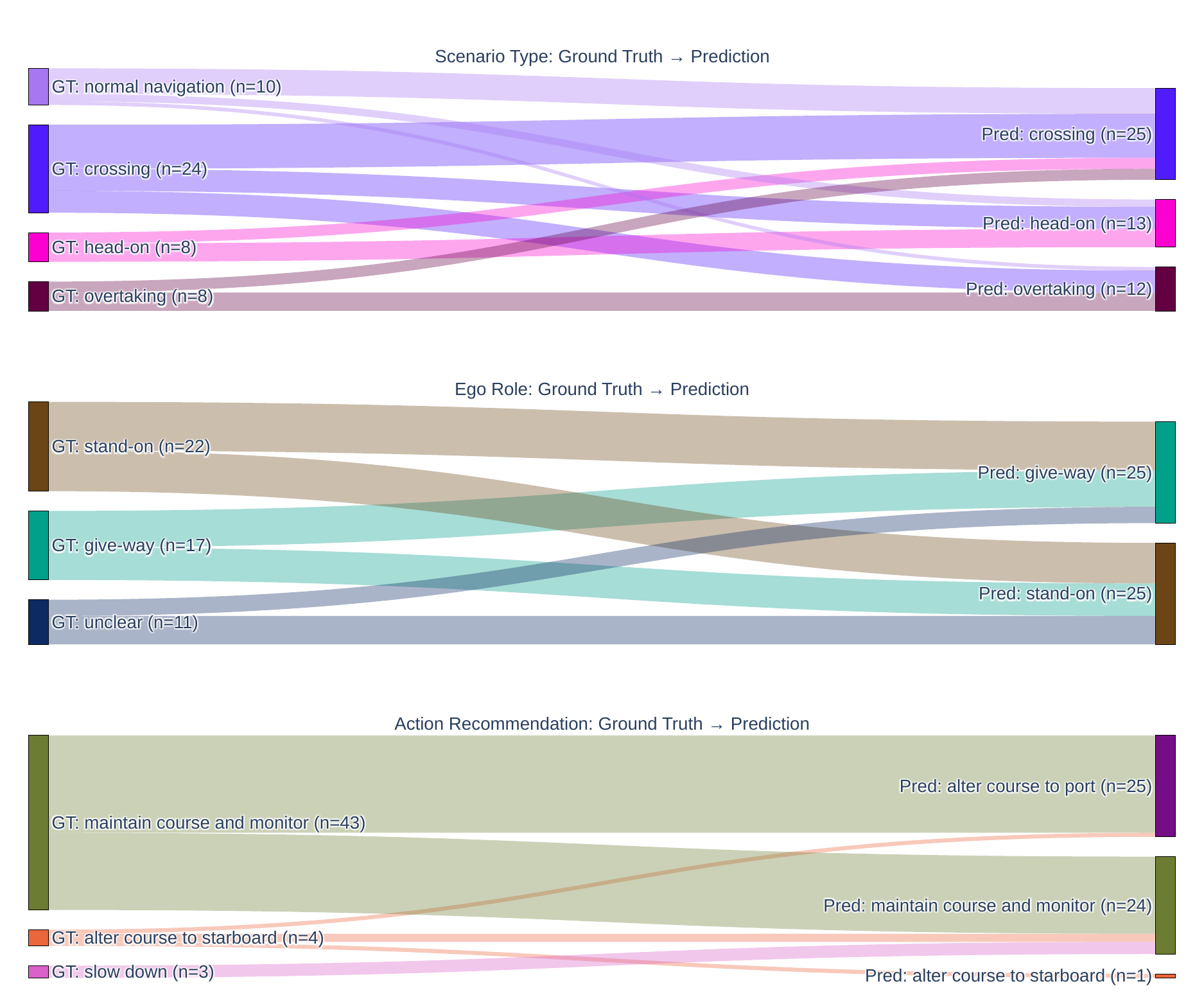}
        \caption{MistralAI Mistral 7B}
        \label{fig:sub2}
    \end{subfigure}
    \caption{Individual scenario evaluations for GPT-4o and Mistral 7B03 using Sankey diagrams.}
    \label{fig:campaigns}
\end{figure*}

Small models continue to default to a single role regardless of scenario. Mistral-7B03, while superficially aligned with the reference, tends to split predictions between stand-on and give-way, often confusing the two. GPT-4o notably fails to identify any reference labels stand-on roles, instead misclassify them as give-way or unclear, as shown in Fig.~\ref{fig:sub1}. In contrast, GPT-5-mini is the only model to consistently match the reference, correctly distributing stand-on, give-way, and unclear roles across scenarios.

\subsection{Action Recommendation}
Action recommendations follow similar trends: Phi4-mini and Gemma3-1b default to a single output, which is often correct, though Phi4-mini incorrectly suggests altering course to port in six cases—contrary to COLREGs. Mistral-7B03 makes this incorrect recommendation in half the scenarios. GPT-4o, despite frequent misclassifications of vessel roles, often recommends the correct starboard maneuver, showing partial rule alignment. GPT-5-mini again aligns closely with the reference, with appropriate starboard recommendations consistent with its scenario classifications.

\subsection{Statistical Evaluation}
To further explore these trends, the association between predicting the Scenario Type and Ego Role / Action Recommendation is evaluated using Cramér's V Score. A V Score of 1 indicates a perfect association, while a V Score of 0 indicates no association. Table~\ref{tab:model_performance} presents the results of this analysis.

\begin{table}[bp]
    \centering
     \caption{Cramér's V Score across LLMs against the reference LLM GPT-5.}
     \label{tab:model_performance}
    \begin{tabular}{lcc}
        \toprule
        \textbf{Model} & \textbf{vs. Ego Role} & \textbf{vs. Action} \\
        \midrule
        Reference GPT-5     & 0.7772 & 0.3397 \\
        \midrule
        OpenAI GPT-5-mini         & 0.3168 & 0.4838 \\
        Google Gemma 3 1B      & --     & --     \\
        OpenAI GPT-4o        & 0.6763 & 0.9125 \\
        MistralAI Mistral 7B    & 0.9895 & 0.6773 \\
        Microsoft Phi 4 Mini   & 1.0000 & 0.9895 \\
        \bottomrule
    \end{tabular}
\end{table}

The reference label shows a strong correlation between scenario type and role (V Score 0.777) and a moderate one between type and action (0.340). GPT-5-mini underestimates the type-role link and slightly overstates the type-action link, despite output patterns resembling the reference. GPT-4o most closely reproduces the reference pattern for the type–role mapping but substantially over-associates the scenario type with action, indicating a stronger internal coupling of scenario interpretation and action choice than what is present in the reference labels. Both Mistral-7B03 and Phi4-mini-it show consistently inflated association strengths across both mappings, suggesting that these smaller models collapse scenario reasoning into near-deterministic patterns. Gemma3-1B cannot be evaluated due to insufficient category variation in its predictions.

\subsection{Individual Evaluation: GPT-4o}
GPT-4o demonstrates several types of errors within this analysis. Most common at the scenario type classification are potential geometric patterns, where head-on scenarios are classified as crossing scenarios, suggesting the model struggles with angular directions. This is consistent with the model misclassifying normal navigation as either overtaking or crossing situations. On the role side, GPT-4o consistently classifies the scenario as give-way instead of stand-on. Furthermore, unclear situations are also labeled as give-way, showing the model tends to be overly cautious. Finally, the action recommendations correlate with the role pattern: as it identities the ego-role as give-way, it recommends altering the course to starboard. This is the correct action according to COLREGs, but not the correct action in the given scenarios. Overall, GPT-4o demonstrates a strong risk-averse bias that, while creating unnecessary inefficiencies due to more frequent maneuvering, does not lead to recommend dangerous or not allowed actions which could put the vessel in danger.

\subsection{Individual Evaluation: Mistral 7B03}
Unlike GPT-4o, Mistral 7B03 struggles significantly with the navigation task. Most notably, the model shows a lack of spatial understanding, frequently mislabeling crossings as overtakes or head-on scenarios, while also flipping head-on situations into crossings. On the ego-role side, Mistral 7B03 shows the opposite behavior to GPT-4o: it misjudges give-way and unclear scenarios as stand-on, leading to potential dangerous action recommendations. Furthermore, on the action side, altering course to port is recommended frequently, an action generally not recommended in the COLREGs (Rules 14, 15). Unlike with GPT-4o, these errors do not seem to be connected, as there is no clear error pattern and errors are contradictory to the information provided to the network. Most notably, there is a clear role-action disconnect, where unsafe action of altering course to port is recommended for a head-on scenario, something that is in clear violation of Rule 14 of the COLREGs. 

\section{Discussion}
These results underscore that vanilla LLMs, online and offline, are not yet suitable for maritime navigation without dedicated fine-tuning. While smaller models like Phi4-mini and Gemma3-1B seemingly rival larger models on pure accuracy metrics, this is largely an artifact of defaulting to deterministic, repetitive outputs, which in case of Mistral 7B03 or Phi4-mini, lead to dangerous stand-on passivity or prohibited, port-turn recommendations. In contrast, the larger models do demonstrate a sophisticated but biased grasp of the COLREGs: GPT-4o exhibits a strong risk-averse tendency, systematically identifying potential conflicts and defaulting to inefficient but compliant starboard maneuvers, while GPT-5-mini shows strong, consistent performance. The high Cramér's V scores of the offline models further confirm that their seemingly correct answers often stem from rigid, over-associated heuristic patterns rather than the more nuanced, situational decision making process evident from the online models.

\section*{DECLARATION OF GENERATIVE AI AND AI-ASSISTED TECHNOLOGIES IN THE WRITING PROCESS}
During the preparation of this work the author(s) used M365 Copilot and OpenAI GPT OSS to improve phrasing of individual sentences and paragraphs. After using this tool/service, the author(s) reviewed and edited the content as needed and take(s) full responsibility for the content of the publication.

\bibliography{ifacconf}

\end{document}